\documentclass[runningheads]{llncs}

\usepackage[T1]{fontenc}
\usepackage{colortbl}
\usepackage{booktabs}
\usepackage{algorithm}            %
\usepackage{algpseudocodex}       %
\usepackage{makecell}
\usepackage{graphicx}
\usepackage{amsmath} 
\usepackage{amsfonts}
\usepackage{microtype}

\usepackage{tikz}
\usepackage{bm}
\usepackage{cite}
\usepackage{siunitx}
\usepackage{amssymb}
\usepackage{adjustbox}
\usepackage{comment}
\usepackage{hyperref}
\usepackage{enumitem}

\usepackage{pgfplots}
\pgfplotsset{compat=1.18}

\usetikzlibrary{positioning, arrows.meta, shapes.geometric, backgrounds, calc, shadows, fit, patterns, decorations.pathmorphing}
\usepackage[table]{xcolor}

\newcommand{\mypar}[1]{\vspace{1pt}\noindent\phantomsection\textbf{#1.}}
\newcommand{\mypartwo}[1]{\vspace{1pt}\noindent\phantomsection\textit{#1.}}

\definecolor{thpmix}{gray}{0.92}
\definecolor{regimecol}{rgb}{0.90,0.94,1.0}
\definecolor{groupband}{gray}{0.92}
\definecolor{nbest}{RGB}{0,102,204}
\definecolor{myblue}{RGB}{60, 120, 216}
\definecolor{mygreen}{RGB}{106, 168, 79}
\definecolor{mypurple}{RGB}{166, 77, 121}
\definecolor{myorange}{RGB}{230, 145, 56}
\definecolor{regimecol}{gray}{0.92}

\DeclareMathOperator*{\argmax}{arg\,max}

\begin{document}

\title{Probabilistic Forecasting of Business Process Executions with Neural Temporal Point Processes}
\titlerunning{Probabilistic Forecasting of Business Process Executions}

\author{Jiaxin Yuan\inst{1} \and
Daniela Grigori\inst{1} \and
Han van der Aa\inst{2}}
\authorrunning{J. Yuan et al.}
\institute{Paris Dauphine University-PSL, Pl. du Mar\'echal de Lattre de Tassigny,
75016 Paris, France\\
\email{jiaxin.yuan@dauphine.eu, daniela.grigori@lamsade.dauphine.fr} \and
University of Vienna, Universit\"atsring 1, 1010 Wien, Austria\\
\email{han.van.der.aa@univie.ac.at}}
\maketitle

\begin{abstract}
Operators of service-based systems act on forecasts of how a running execution
will continue, and such a forecast is actionable only if its reliability is
known. Mainstream deep-learning models for this task are discriminative and
deterministic: they emit a single next activity and a single remaining-time
estimate, without a distribution to reason over. We instead cast the problem as
generative sequence modelling with marked temporal point processes, which
define a joint density over the next mark and its inter-event time and
therefore deliver predictive distributions by construction. Real event logs
violate the simple-point-process assumption these models rest on, since
consecutive events frequently carry identical timestamps; we handle such ties
explicitly and combine a transformer encoder with a mixture decoder over
inter-event times, trained by exact log-likelihood. On ten public logs, the
resulting model matches discriminative baselines on point accuracy, dominates
them on the calibration and sharpness of remaining-time distributions, and is
the cheapest at inference, since a full predictive distribution is obtained in
a single forward pass without sampling.

\keywords{Event sequence modelling \and Marked temporal point processes \and
Probabilistic forecasting \and Calibration}
\end{abstract}

\section{Introduction}
\label{sec:intro}

Modern organizations frequently realize their processes as
orchestrations of software and human services, whose execution is recorded as
event data~\cite{van2016data}. Anticipating how an ongoing \emph{case}---a
single, running execution of such a process---will unfold turns passive
monitoring into proactive control: it lets operators reallocate resources,
escalate at-risk cases, and intervene before a case's \emph{quality of service}
(QoS) degrades or its \emph{service-level agreement} (SLA) is
violated~\cite{leitner2010runtime,metzger2015ComparingAC}. \emph{Predictive
process monitoring} (PPM) addresses this need, forecasting the future of running
cases from historical event logs~\cite{rama2021deep}. Its two central
time-oriented tasks are \emph{next event prediction}, forecasting the type and
time of the next event, and \emph{remaining time prediction}, forecasting the
time until a case completes.

For acting on such a forecast at runtime, however, a prediction is only as valuable
as the confidence attached to it. The dominant paradigm in PPM offers no such
confidence: state-of-the-art deep-learning models---whether sequence models such as LSTMs and Transformers~\cite{rama2021deep,bukhsh2021processtransformer},
or graph neural networks that encode control flow~\cite{amiri2024pgtnet,weinzierl2021exploring}---are almost
exclusively \emph{discriminative} and \emph{deterministic}, optimized directly for point predictions on downstream tasks.
They return a single activity or timepoint but say little
about how (un)certain it is, even though this uncertainty is exactly what runtime
QoS and SLA management depend on: a case that will \emph{probably} miss its
deadline warrants a different intervention than one that almost certainly will.
Underlying this limitation is that such models treat time as an auxiliary feature rather
than a first-class prediction target.

Temporal point processes (TPPs) offer a principled way to close this gap. As
probabilistic \emph{generative} models of event sequences that unfold in continuous
time~\cite{daley2008introduction,shchur2021neural}, they natively produce a predictive
\emph{distribution} over both the next activity and its timing---and, by extension,
over a case's remaining duration. This makes
them a natural fit for uncertainty-aware service and process monitoring. Their use
in PPM, however, has barely been explored, for a concrete reason: most TPPs,
including the widely used class of Hawkes processes~\cite{hawkes1971point}, assume a \emph{simple} point
process in which no two events may occur at the same moment. Real event logs
frequently violate this assumption, as they routinely record consecutive events with
\emph{equal timestamps} (up to 46\% in the logs we study), owing to limited timestamp resolution~\cite{fischer2020enhancing},
batching~\cite{martin2015batch}, or parallel execution~\cite{suriadi2017event}. 
Standard TPPs are misspecified for such logs: their predictive
distributions place zero probability on the very transitions the data contains,
forfeiting the distributional benefits that motivate their use.

Therefore, we propose \textit{MoTPP}, an adaptation of the
Transformer Hawkes Process (THP)~\cite{zuo2020transformer} that brings the generative TPP paradigm to
multi-task PPM. Inspired by the intensity-free formulation~\cite{Shchur2020Intensity}, we replace THP's conditional intensity and its time prediction head with a \emph{mixture time head} that can place probability on equal timestamps. We train it jointly with a next activity head, so that the model predicts the next
activity and its time together. From the resulting generative model, we then obtain
an uncertainty-aware remaining time distribution through Monte Carlo rollout.
{MoTPP} is thus the first generative temporal point
process for multi-task PPM that natively handles the equal timestamps that are pervasive in real event logs.
 
We evaluate \textit{MoTPP} on ten real-world event logs against
representative state-of-the-art PPM methods, focusing not on point accuracy alone
but also on what matters for runtime monitoring---calibrated uncertainty and
low-latency inference---and observe that:
\begin{itemize}[topsep=0pt]
  \item under conventional point metrics, \textit{MoTPP} is
    competitive, though not dominant: it attains
    the lowest next event time error on high-frequency, small-vocabulary logs, whereas state-of-the-art sequence models retain an edge in next activity accuracy on larger,
    lower-frequency logs;
  \item evaluated as a probabilistic predictor, \textit{MoTPP} yields the
    best-calibrated remaining time predictions among all evaluated methods---the
    property most directly relevant to proactive SLA and QoS management; and
  \item \textit{MoTPP} attains the lowest overall training cost and the
    fastest inference on all ten logs, scoring each prefix’s next event prediction in a single
    non-autoregressive forward pass, which makes it attractive for online,
    low-latency runtime monitoring.
\end{itemize}

\noindent In the remainder, \autoref{sec:pre} introduces
preliminaries and \autoref{sec:motivation} motivates our work.
\autoref{sec:model} presents the \textit{MoTPP} model, \autoref{sec:evaluation}
reports our evaluation, \autoref{sec:rw} discusses related work, and
\autoref{sec:conclusion} concludes.

\section{Preliminaries}
\label{sec:pre}
This section defines the notation and prediction tasks used throughout the paper.
 
\mypar{Event data}
A trace $\sigma=\langle e_1,\dots,e_{|\sigma|}\rangle$ is a sequence of events of
a case, and an event log $L$ is a collection of traces of the same process. Each
event $e_i$ is a tuple $(a_i,c,t_i)$ with activity $a_i$, case id $c$, and
timestamp $t_i$; we write $\tau_i=t_i-t_{i-1}$ (for $i>1$, with $\tau_1=0$) for the
inter-event time. A \emph{prefix} $\mathcal{H}_k(\sigma)=\langle e_1,\dots,e_k\rangle$,
$1 \le k <|\sigma|$, captures the execution of a case up to its $k$-th event.
 
\mypar{Prediction tasks}
Given a prefix $\mathcal{H}_k(\sigma)$, \emph{next event prediction} forecasts the next
activity and the next event timestamp, whereas \emph{remaining time prediction}
forecasts the time until the case completes. We capture these with three
predictors that each map a prefix to a prediction: $\hat a$ for the next activity
$a_{k+1}$, $\hat t$ for the next time $t_{k+1}$, and $\hat r$ for the
remaining time until completion. Formally,
\begin{equation*}
  \hat a(\mathcal{H}_k(\sigma))\approx a_{k+1},\qquad
  \hat t(\mathcal{H}_k(\sigma))\approx t_k + \tau_{k+1},\qquad
  \hat r(\mathcal{H}_k(\sigma))\approx t_{|\sigma|}-t_k .
\end{equation*}

\section{Motivation}
\label{sec:motivation}
This section introduces temporal point processes (TPPs), explains why they suit
PPM, and shows how the limitations of a state-of-the-art TPP model motivate our
work.

\mypar{Temporal point processes}
TPPs provide a probabilistic generative framework for sequences of discrete
events in continuous time~\cite{daley2008introduction}. Conditioned on the
history of past events, a TPP defines a joint distribution over when the next
event occurs and which type it carries\footnote{A TPP whose events carry types is
formally a marked TPP with a finite mark space~\cite{daley2008introduction};
following THP, we refer to marks as event types~\cite{zuo2020transformer}.},
typically parameterized through a conditional intensity function $\lambda^*$ or a
conditional density $p^*$. Classical models such as the Hawkes
process~\cite{hawkes1971point} use handcrafted parametric intensities, whereas
recent neural TPPs replace these with deep neural
networks~\cite{du2016recurrent,mei2017neural,zuo2020transformer}.

\mypar{Suitability for PPM}
Three properties make TPPs a natural fit for PPM. First, the modeled object
matches: a TPP models event times in the continuous domain $\mathbb{R}^+$, so the
traces in an event log are realizations of an underlying TPP and an activity's
occurrence time becomes a prediction target in its own right. Second, being
generative, a TPP yields a full predictive distribution rather than a point
prediction, providing the calibrated uncertainty that runtime QoS and SLA
management require~\cite{shchur2021neural,ghahramani2015probabilistic}. Third, TPPs are multi-task by construction: a single one-step conditional
model supports next activity and next-time prediction directly, and yields
remaining time prediction by sampling continuations of the running case, without
task-specific decoders.

\begin{figure}[htbp]
\centering
\vspace{-2em}
\begin{adjustbox}{width=1\textwidth}
\definecolor{colInput}{RGB}{174, 214, 241}%
\definecolor{colTempo}{RGB}{169, 223, 191}%
\definecolor{colEnc}{RGB}{209, 196, 233}%
\definecolor{colNorm}{RGB}{230, 224, 243}%
\definecolor{colLSTM}{RGB}{174, 198, 231}%
\definecolor{colState}{RGB}{132, 186, 225}%
\definecolor{darkLine}{RGB}{80, 80, 80}%
\begin{tikzpicture}[
    font=\sffamily\normalsize,
    node distance=0.45cm and 0.32cm,
    styleBox/.style={
        rectangle, rounded corners=3pt, draw=darkLine, line width=1pt,
        align=center, font=\normalsize, text=black, inner sep=4pt,
        drop shadow={opacity=0.15, shadow xshift=1pt, shadow yshift=-1pt},
        minimum height=1.0cm, minimum width=1.3cm
    },
    styleEnc/.style={styleBox, minimum height=1.75cm},
    styleCircle/.style={
        circle, draw=darkLine, line width=1pt, font=\normalsize, text=black, inner sep=1pt,
        drop shadow={opacity=0.15, shadow xshift=1pt, shadow yshift=-1pt},
        minimum size=0.8cm
    },
    stylePlusNode/.style={
        circle, draw=darkLine, line width=1pt, minimum size=0.62cm,
        drop shadow={opacity=0.15, shadow xshift=1pt, shadow yshift=-1pt},
        path picture={
            \draw[darkLine, thick]
                (path picture bounding box.north) -- (path picture bounding box.south)
                (path picture bounding box.west) -- (path picture bounding box.east);
        }
    },
    styleArrow/.style={-Stealth, thick, draw=darkLine},
    styleRes/.style={-Stealth, thick, draw=darkLine!70},
    styleLabel/.style={font=\normalsize, text=black, align=center},
]

\node[styleBox, fill=colInput, minimum width=1.9cm] (evseq) {Event\\ sequence};
\node[styleBox, fill=colInput, right=0.45cm of evseq, minimum width=1.9cm] (input_emb) {Input\\Embedding};
\node[stylePlusNode, right=0.5cm of input_emb] (plus_op) {};

\node[styleCircle, fill=colTempo, above=0.9cm of plus_op] (te) {TE};
\node[right=0.15cm of te, styleLabel, font=\normalsize] (te_lab) {Temporal encoding};

\draw[styleArrow] (evseq) -- (input_emb);
\draw[styleArrow] (input_emb) -- (plus_op);
\draw[styleArrow] (te) -- (plus_op);
\draw[styleArrow, rounded corners=3pt] (evseq.north) |- (te.west);

\node[styleEnc, fill=colEnc,  right=0.75cm of plus_op, minimum width=1.7cm] (mha)   {Multi-\\Head\\Attention};
\node[styleEnc, fill=colNorm, right=0.3cm of mha, minimum width=1.3cm]      (norm1) {Add. \&\\Norm.};
\node[styleEnc, fill=colEnc,  right=0.3cm of norm1, minimum width=1.45cm]   (ff)    {Feed\\Forward};
\node[styleEnc, fill=colNorm, right=0.3cm of ff, minimum width=1.3cm]       (norm2) {Add. \&\\Norm.};

\draw[styleArrow] (plus_op.east) -- (mha.west);
\draw[styleArrow] (plus_op.east) -- ([yshift=0.5cm]mha.west);
\draw[styleArrow] (plus_op.east) -- ([yshift=-0.5cm]mha.west);
\draw[styleArrow] (mha) -- (norm1);
\draw[styleArrow] (norm1) -- (ff);
\draw[styleArrow] (ff) -- (norm2);

\coordinate (res1) at ($(mha.west)+(-0.45cm,0)$);
\draw[styleRes] (res1) -- ++(0,1.2cm) -| (norm1.north);
\coordinate (res2) at ($(ff.west)+(-0.2cm,0)$);
\draw[styleRes] (res2) -- ++(0,1.2cm) -| (norm2.north);

\node[
    draw=darkLine!65, dashed, thick, rounded corners=6pt,
    fit=(mha)(norm2)(res1)(res2), inner sep=5pt
] (encoder_block) {};
\node[above=-0.14cm of encoder_block.south, font=\normalsize\bfseries, text=black] {$N\times$};

\node[styleBox, fill=colLSTM, draw=darkLine, dashed, right=0.5cm of encoder_block, minimum height=1.15cm, minimum width=1.4cm] (lstm) {LSTM};
\node[below=0.08cm of lstm, styleLabel, font=\small\itshape] {optional};
\node[styleBox, fill=colState, right=0.42cm of lstm, minimum width=1.25cm] (h) {$\mathbf{h}(t_k)$};
\draw[styleArrow] (encoder_block.east) -- (lstm.west);
\draw[styleArrow] (lstm.east) -- (h.west);

\end{tikzpicture}
 \end{adjustbox}
\caption{The THP-backbone, mapping an event sequence to the hidden state \(\mathbf h(t_k)\).}
\vspace{-2em}
\label{fig:thp-backbone}
\end{figure}
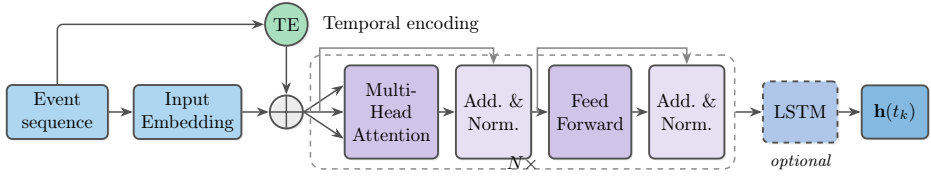

\mypar{Transformer Hawkes Process}
Despite this fit, neural TPPs remain largely unexplored for predictive
monitoring, with only a first attempt in this direction~\cite{nguyen2024temporal}.
A state-of-the-art neural TPP is the Transformer Hawkes Process
(THP)~\cite{zuo2020transformer}, which takes an \textit{event sequence} of
(timestamp, event-type) pairs and, through $N$ causally masked self-attention
layers followed by an LSTM (an optional configuration provided in the official THP codebase), produces a hidden state $\mathbf h(t_k)$ that
summarizes the prefix $\mathcal H_k(\sigma)$. We refer to this encoder, shown
in~\autoref{fig:thp-backbone}, as the \textit{THP-backbone}. Its output layer,
however, assigns zero probability to equal timestamps, so that any prefix
containing simultaneous events receives zero likelihood---even though such
timestamps are pervasive in event logs. This misspecification motivates the
\textit{MoTPP} model presented next.

\section{MoTPP}
\label{sec:model}
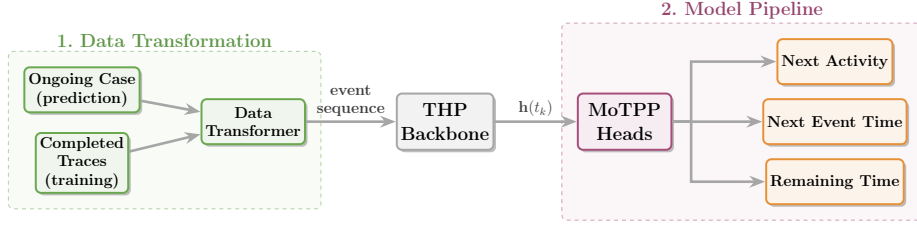
\begin{figure}[!h]
    \centering
    \resizebox{\textwidth}{!}{
\begin{tikzpicture}[
    font=\bfseries\large,
    node distance=1.5cm and 2cm,
    data/.style={
        rectangle,
        rounded corners,
        minimum width=2cm,
        minimum height=1.2cm,
        text centered,
        align=center,
        draw=mygreen,
        fill=mygreen!10,
        line width=1.5pt,
        drop shadow
    },
    model/.style={
        rectangle,
        rounded corners,
        minimum width=2.5cm,
        minimum height=1.5cm,
        text centered,
        align=center,
        draw=mypurple,
        fill=mypurple!10,
        line width=1.5pt,
        font=\bfseries\Large,
        drop shadow
    },
    output/.style={
        rectangle,
        rounded corners,
        minimum width=2.5cm,
        minimum height=1.2cm,
        text centered,
        align=center,
        draw=myorange,
        fill=myorange!10,
        line width=1.5pt,
        font=\bfseries\large,
        drop shadow
    },
    arrow/.style={
        ->,
        >=Stealth,
        line width=2pt,
        draw=gray!70
    },
    label_text/.style={
        font=\bfseries\large,
        text=darkgray,
        align=center
    },
    group_box/.style={
        draw=#1!50,
        dashed,
        thick,
        rounded corners,
        inner sep=0.4cm,
        fill=#1!5
    }
]

\node[data] (case) {Ongoing Case\\(prediction)};
\node[data, below=0.5cm of case] (el) {Completed\\ Traces\\(training)};
\node[data, right=1.6cm of case, yshift=-0.85cm] (pref) {Data\\Transformer};

\draw[arrow] (case) -- (pref);
\draw[arrow] (el) -- (pref);

\node[model, draw=gray!70, fill=gray!10, right=2.4cm of pref] (thp) {THP\\Backbone};
\node[model, right=2.2cm of thp] (heads) {MoTPP\\Heads};

\draw[arrow] (pref) -- node[above, label_text] {event\\sequence} (thp);
\draw[arrow] (thp) -- node[above, label_text] {$\mathbf{h}(t_k)$} (heads);

\node[output, right=2.4cm of heads] (pred_time) {Next Event Time};
\node[output, above=0.35cm of pred_time] (pred_act) {Next Activity};
\node[output, below=0.35cm of pred_time] (pred_rem) {Remaining Time};

\draw[arrow] (heads.east) -- ++(0.5,0) |- (pred_act.west);
\draw[arrow] (heads.east) -- (pred_time.west);
\draw[arrow] (heads.east) -- ++(0.5,0) |- (pred_rem.west);

\begin{scope}[on background layer]
    \node[group_box=mygreen, fit=(case) (el) (pref),
          label={[text=mygreen!90!black, font=\bfseries \Large]above:1. Data Transformation}] (box_trans) {};

    \node[group_box=mypurple, fit=(heads) (pred_act) (pred_time) (pred_rem),
          label={[text=mypurple!90!black, font=\bfseries \Large]above:2. Model Pipeline}] (box_model) {};
\end{scope}

\end{tikzpicture}
     }
\caption{Overview of MoTPP for multi-task predictions in PPM}

\label{fig:workflow}
\end{figure}

We propose \textit{MoTPP}, a probabilistic generative model for multi-task
PPM, with an overview shown in \autoref{fig:workflow}. Given a prefix of a running case, MoTPP
predicts a joint distribution over the next activity, next event time, and remaining
time. Because each prediction is derived from this distribution rather than produced
as a point prediction, the model is uncertainty-aware, yielding a
predictive distribution over the remaining time.
As its name suggests, it keeps the THP-backbone as its sequence encoder but
replaces THP's intensity-based output with a zero-inflated LogNormal mixture
density, which---unlike the original---places probability mass on the equal timestamp pervasive in event logs.
 
\autoref{sec:datatransformation} describes how event data is cast into the format the model
consumes---prefixes drawn from completed traces at training time and a prefix of an ongoing case at inference time, while \autoref{sec:train-inf} presents the model itself: its architecture, training objective, and inference procedure.

\subsection{Data Transformation}
\label{sec:datatransformation}
As motivated in~\autoref{sec:motivation}, a completed trace $\sigma$ can be
regarded as a realization of a TPP, and a prefix $\mathcal{H}_k(\sigma)$ as a
partially observed realization---the history upon which a TPP is conditioned up to timestamp $t_k$. This
section makes that correspondence operational at two levels, illustrated
in~\autoref{fig:workflow}. At the \emph{event level}, a projection casts each
event into the pair the model consumes. At the \emph{trace level}, three
mismatches remain between event log traces and TPP realizations---two resolved here by restructuring the data, one deferred to the
model.

\mypar{Projection}
Each event $e_i$ in $\mathcal{H}_k(\sigma)$ is projected onto the pair
$(a_i, t_i)$: the activity $a_i$ serves as the event type, and $t_i$
is the elapsed time since the first event of the case.

\mypar{Termination}
Cases terminate, whereas TPP realizations are unbounded. Since the TPP carries no explicit termination symbol, we resolve this at
inference (\autoref{sec:train-inf}). The activity set $\mathcal{A}$ is simply the
activities occurring in the log.

\mypar{Case separation}
An event log interleaves many concurrent cases, whereas a TPP realization is a
single sequence. We therefore treat each case as its own realization operating under its own
clock, never concatenating cases into a global stream.

\mypar{Equal timestamps}
Unlike the previous two, this mismatch cannot be removed by restructuring the
data. Equal timestamps violate the \textit{simple} point-process assumption that
no two events occur at the same moment; the model itself accounts for them
(\autoref{sec:train-inf}).

\subsection{Model Pipeline} \label{sec:train-inf}
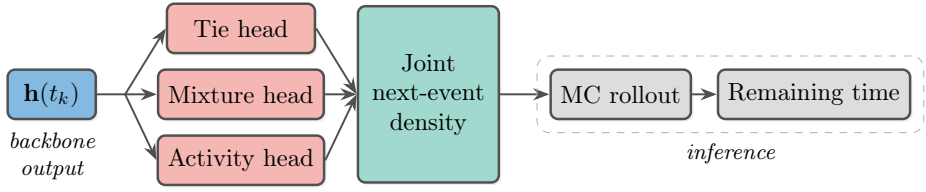
\begin{figure}[htbp]
\centering

\begin{adjustbox}{width=1\textwidth}
\definecolor{colState}{RGB}{132, 186, 225}%
\definecolor{colHead}{RGB}{245, 183, 177}%
\definecolor{colDens}{RGB}{162, 217, 206}%
\definecolor{colInfer}{RGB}{221, 221, 221}%
\definecolor{darkLine}{RGB}{80, 80, 80}%
\begin{tikzpicture}[
    font=\sffamily\normalsize,
    node distance=0.45cm and 0.32cm,
    styleBox/.style={
        rectangle, rounded corners=3pt, draw=darkLine, line width=1pt,
        align=center, font=\normalsize, text=black, inner sep=4pt,
        drop shadow={opacity=0.15, shadow xshift=1pt, shadow yshift=-1pt},
        minimum height=0.7cm, minimum width=1.2cm
    },
    styleArrow/.style={-Stealth, thick, draw=darkLine},
    styleLabel/.style={font=\normalsize, text=black, align=center},
]

\node[styleBox, fill=colState, minimum width=1.25cm] (h) {$\mathbf{h}(t_k)$};
\node[styleLabel, below=0.06cm of h, font=\small\itshape] {backbone\\output};

\node[styleBox, fill=colHead, right=0.85cm of h, minimum width=2.1cm] (time_head) {Mixture head};
\node[styleBox, fill=colHead, above=0.2cm of time_head, minimum width=2.1cm] (tie_head)  {Tie head};
\node[styleBox, fill=colHead, below=0.2cm of time_head, minimum width=2.1cm] (mark_head) {Activity head};

\coordinate (fan) at ($(time_head.west)+(-0.45cm,0)$);
\draw[thick, draw=darkLine] (h.east) -- (fan);
\draw[styleArrow] (fan) -- (tie_head.west);
\draw[styleArrow] (fan) -- (time_head.west);
\draw[styleArrow] (fan) -- (mark_head.west);

\node[styleBox, fill=colDens, right=0.45cm of time_head,
      minimum width=2.0cm, minimum height=2.5cm] (density) {Joint\\next-event\\density};

\draw[styleArrow] (tie_head.east)  -- (density.west);
\draw[styleArrow] (time_head.east) -- (density.west);
\draw[styleArrow] (mark_head.east) -- (density.west);

\node[styleBox, fill=colInfer, right=0.7cm of density, minimum width=1.8cm] (rollout) {MC rollout};
\node[styleBox, fill=colInfer, right=0.35cm of rollout, minimum width=2.3cm]  (rt) {Remaining time};
\draw[styleArrow] (density.east) -- (rollout.west);
\draw[styleArrow] (rollout.east) -- (rt.west);

\node[draw=darkLine!45, dashed, rounded corners=5pt, fit=(rollout)(rt), inner sep=5pt] (infbox) {};
\node[below=0.03cm of infbox.south, font=\small\itshape, text=black] {inference};

\end{tikzpicture}
 \end{adjustbox}

\caption{Prediction heads and inference procedure of MoTPP.}
\label{fig:architecture}

\end{figure}

After transformation, we now present the model itself, following the pipeline shown in \autoref{fig:architecture}: its architecture, which equips the THP-backbone with new prediction heads; its training objective, which fits the resulting conditional density of the next event by maximum likelihood; and its inference procedure, which decodes the PPM tasks from this density.

\mypar{Architecture} \label{sec:architecture}
We keep the THP-backbone and replace its output layer with three linear heads that jointly parameterize a conditional density over the next event, rather than emitting point predictions.

\mypartwo{Prediction heads} 
Unlike THP's single event type and time heads, we apply three linear heads to \(\mathbf{h}(t_k)\in\mathbb{R}^{M}\):

\begin{enumerate}
    \item a tie head \(\mathbf{w}^{\pi}\in\mathbb{R}^{M}\) producing a zero-inflation logit
    \item a mixture head \(\mathbf{W}^{\text{m}}\in\mathbb{R}^{3K\times M}\) with \(K\) representing the number of LogNormal components; and
    \item an activity head \(\mathbf{W}^{\text{a}}\in\mathbb{R}^{|\mathcal{A}|\times M}\) producing next activity logits over \(\mathcal{A}\)
\end{enumerate}
\noindent
Writing \(\tau := \tau_{k+1}\), the tie head gives the logit of \(\pi\), the probability of an instantaneous transition (\(\tau=0\)); the mixture head parameterizes the density of \(\tau\) given \(\mathbf{h}(t_k)\), with \(K\) components capturing the multimodal inter-event times of event logs (e.g.\ fast automated handoffs alongside slow human steps) that a single LogNormal cannot. Together the three heads define the conditional joint density over the next activity and its time (\autoref{eq:joint}), learned by maximum likelihood.

\mypar{Training} \label{sec:train}
We model the conditional joint density of the next activity and time given the prefix \(\mathcal{H}_k(\sigma)\), using the asterisk $(*)$ to denote this conditioning by convention:
\begin{equation}
p^*(a,\tau) = p^*(a)\,p^*(\tau).
\label{eq:joint}
\end{equation}
This factorization assumes activity and time are conditionally independent given the prefix, a standard simplification in neural TPPs~\cite{shchur2021neural}; their dependence is captured only through the shared representation \(\mathbf{h}(t_k)\).

\mypartwo{Activity density} The next activity follows a categorical distribution over \(\mathcal{A}\),
\begin{equation}
p^*(a) = \mathrm{softmax}\!\big(\mathbf{W}^{\text{a}}\mathbf{h}(t_k)\big)_a ,
\end{equation}
where the subscript $a$ extracts the entry corresponding to activity $a$ from the resulting probability vector.

\mypartwo{Zero-inflated LogNormal mixture} We model \(\tau\) with a zero-inflated mixture: with probability \(\pi\) the next event is instantaneous (\(\tau=0\)), and with probability \(1-\pi\), the time is drawn from a \(K\)-component LogNormal mixture,
\begin{equation}
p^*(\tau) = \pi\,\delta_0(\tau) + (1-\pi)\sum_{m=1}^{K} w_m \,\mathrm{LN}\!\left(\tau;\mu_m,s_m\right),
\label{eq:cond-density}
\end{equation}
where \(\delta_0\) denotes the point mass at \(0\), \(m\) indexes the mixture components with weights \(w_m\geq 0\), \(\sum_{m=1}^{K} w_m=1\), and \(\mathrm{LN}(\tau;\mu_m,s_m)=\frac{1}{\tau\,s_m\sqrt{2\pi}}\exp\!\big(-\tfrac{(\ln \tau-\mu_m)^2}{2s_m^2}\big)\) is the LogNormal density with \(\mu_m\) and \(s_m\) the mean and standard deviation of \(\ln\tau\) under component \(m\). The parameters are derived from the heads at position \(k\): the zero-inflation probability \(\pi=\mathrm{sigmoid}(\mathbf{w}^{\pi\top}\mathbf{h}(t_k))\), the mixture weights \(\mathbf{w}=\mathrm{softmax}(\cdot)\), the log-scale means \(\boldsymbol{\mu}\), and the positive scales \(\mathbf{s}=\mathrm{softplus}(\cdot)+\varepsilon\), where \((\mathbf{w},\boldsymbol{\mu},\mathbf{s})\) jointly constitute the \(3K\) outputs of \(\mathbf{W}^{\text{m}}\mathbf{h}(t_k)\).

\mypartwo{Maximum likelihood estimation} Each prediction position contributes a time term and an activity term. The negative log-likelihood of the inter-event time is
\begin{equation}
\ell_{\tau} =
\begin{cases}
-\log \pi, & \tau = 0,\\[2pt]
-\log(1-\pi) - \log \sum_{m=1}^{K} w_m\,\mathrm{LN}(\tau;\mu_m,s_m), & \tau > 0,
\end{cases}
\end{equation}
and the activity term is defined as the cross-entropy loss \(\ell_{a}=-\log p^*(a_{k+1})\). The training objective sums both terms over all prediction positions of the \(N\) training cases,
\begin{equation}
    \min_{\theta}\; \sum_{i=1}^{N} \sum_{k=1}^{|\sigma_i|-1} \big(\ell_{\tau}^{(i,k)} + \ell_{a}^{(i,k)}\big),
\end{equation}
where \(\theta\) encompasses all learnable parameters of the backbone and the three heads.

\mypar{Inference} \label{sec:inference}
Given a prefix \(\mathcal{H}_k(\sigma)\) (an ongoing case), the next activity predictor is the mode of the categorical distribution,
\begin{equation}
\hat a(\mathcal{H}_k(\sigma)) = \argmax_{a \in \mathcal{A}}\, p^*(a),
\end{equation}
and the next event time predictor is the closed-form expectation of~\autoref{eq:cond-density},
\begin{equation}
\hat t(\mathcal{H}_k(\sigma)) = \mathbb{E}[\tau \mid \mathcal{H}_k(\sigma)] = (1-\pi)\sum_{m=1}^{K} w_m \exp\!\Big(\mu_m + \tfrac{1}{2}s_m^2\Big),
\end{equation}
recovering the next timestamp as $t_k + \hat t(\mathcal{H}_k(\sigma))$. The remaining time predictor accumulates these one-step expectations across the suffix, conditioning on the true prefix at each step:
\begin{equation}
\hat r(\mathcal{H}_k(\sigma)) = \sum_{j=k}^{|\sigma|-1} \mathbb{E}\!\left[\tau \mid \mathcal{H}_j(\sigma)\right],
\label{eq:rt-point}
\end{equation}
where the ground truth supplies both the horizon $J = \vert{}\sigma\vert{} - k$ and the intermediate events, since $\mathcal{A}$ lacks an end-of-case symbol. Consequently, \autoref{eq:rt-point} is a teacher-forced plug-in predictor rather than the model's true generative expectation. For an uncertainty-aware prediction, we drop this conditioning: we ancestrally sample $S$ continuations over the same horizon $J$, and their accumulated times form an empirical remaining time distribution summarized by its sample mean and standard deviation.

\section{Evaluation}
\label{sec:evaluation}

We evaluate MoTPP along three dimensions: point predictive accuracy, uncertainty quality, and computational efficiency. We first describe the datasets, baselines, metrics, and experimental protocol, and then discuss the obtained results.%

\subsection{Experimental Setup}
\label{sec:eval:setup}

\mypar{Datasets}
We use ten public event logs spanning clinical, incident and request management, financial, and travel-reimbursement processes; \autoref{tab:datasets} reports their statistics after preprocessing. To relate model behavior to log characteristics, we cluster the logs with Ward's method~\cite{ward1963hierarchical} on the two statistics that drive TPP modeling difficulty, the median inter-event time and the number of distinct activities (shaded in~\autoref{tab:datasets}). This yields a high-frequency, small-vocabulary group (Group~A), a low-frequency, larger-vocabulary group (Group~B), and one very-high-frequency, large-vocabulary outlier (\textsc{BPIC15-1}).

\begin{table}[!h]
\centering
\caption{\textsc{Dataset statistics.} Cases, Events and Act: numbers of cases, events, and distinct activities; Avg.\,len: mean case length; Med.\,$\Delta t$: median inter-event time; Avg.\,dur\,(d): mean case duration in days. Shaded columns (Act, Med.\,$\Delta t$) are the cluster-defining statistics.}

\label{tab:datasets}
\setlength{\tabcolsep}{4pt}
\renewcommand{\arraystretch}{1.15}
\begin{tabular}{@{}l rr >{\columncolor{regimecol}}r r >{\columncolor{regimecol}}r r@{}}
\toprule
\textbf{Log} & \textbf{Cases} & \textbf{Events} & \textbf{Act} & \textbf{Avg.\,len} & \textbf{Med.\,$\Delta t$} & \textbf{Avg.\,dur\,(d)} \\
\midrule
\rowcolor{groupband}\multicolumn{7}{@{}l@{}}{\textbf{Group A} --- high-frequency, small-vocab} \\
\textsc{Sepsis}    & 1\,049  & 15\,214  & 16  & 14.5 & 6.5\,min  & 28.5 \\
\textsc{BPIC13I}   & 7\,554  & 65\,533  & 4   & 8.7  & 10.1\,min & 12.1 \\
\textsc{BPIC12}    & 13\,087 & 262\,200 & 24  & 20.0 & 48\,s     & 8.6  \\
\addlinespace[1.5pt]
\rowcolor{groupband}\multicolumn{7}{@{}l@{}}{\textbf{Group B} --- low-frequency, relatively large-vocab} \\
\textsc{HelpDesk}  & 4\,580  & 21\,348  & 14  & 4.7  & 3.28\,d   & 40.9 \\
\textsc{BPIC20DD}  & 10\,500 & 56\,437  & 17  & 5.4  & 1.02\,d   & 11.5 \\
\textsc{BPIC20RFP} & 6\,886  & 36\,796  & 19  & 5.3  & 1.11\,d   & 12.0 \\
\textsc{BPIC20PTC} & 2\,099  & 18\,246  & 29  & 8.7  & 22.5\,h   & 36.8 \\
\textsc{BPIC20ID}  & 6\,449  & 72\,151  & 34  & 11.2 & 2.00\,d   & 86.5 \\
\textsc{BPIC20TPD} & 7\,065  & 86\,581  & 51  & 12.3 & 1.74\,d   & 87.4 \\
\addlinespace[1.5pt]
\rowcolor{groupband}\multicolumn{7}{@{}l@{}}{\textbf{Outlier} --- very-high-frequency, large-vocab} \\
\textsc{BPIC15-1}  & 1\,199  & 52\,217  & 398 & 43.6 & 1\,s      & 95.7 \\
\bottomrule
\end{tabular}

\end{table} 
\mypar{Preprocessing}
All event logs are preprocessed following the strict temporal-splitting approach~\cite{weytjens2021creating}, via the following steps:

\mypartwo{Duplicate removal and case-duration trimming} We drop all fully duplicated events ($712$ across the ten datasets). Rather than specifying a fixed maximum case duration as in the original approach, we trim the longest $5\%$ of cases, yielding a data-dependent threshold that adapts consistently across all datasets.

\mypartwo{End-of-dataset debiasing} We drop cases that start too late to plausibly complete within the observed window, i.e., those whose start time exceeds the final log timestamp minus the longest retained case duration ($3{,}274$ cases).\footnote{For BPIC13I, case arrivals concentrate near the end of the log, so this step would leave too few training cases; it is skipped for this dataset.}

\mypartwo{Strict temporal split} The most recent $20\%$ of cases by start time form the test set; the training set retains only those cases that complete before the earliest test-case start time.

\mypar{Compared methods}
We compare MoTPP with models from four families.

\mypartwo{Neural TPP (THP-B, THP-M)} Two variants of THP~\cite{zuo2020transformer} sharing the same backbone and next activity head: \emph{THP-B} (Baseline) with the original point-regression time head, which enters only the point-prediction comparisons; and our proposed \emph{THP-M} (Mixture) with the zero-inflated LogNormal time head for probabilistic prediction.

\mypartwo{State-of-the-art multi-task PPM models (SuTraN, ED-LSTM)} \emph{SuTraN}~\cite{wuyts2024sutran}, a Transformer encoder-decoder, and \emph{ED-LSTM}~\cite{taymouri2021deep,ketyko2022averages,wuyts2024sutran}, an encoder-decoder LSTM, both designed for multi-task prediction.

\mypartwo{Classical TPP (H-uni, H-mk)} Two variants of an exponential-kernel Hawkes process~\cite{hawkes1971point} fitted by maximum likelihood: a univariate model on inter-event times only (\emph{H-uni}), and a marked model with one intensity dimension per activity (\emph{H-mk}).

\mypartwo{Uncertainty-aware models (LA-CR$\star$, UQ$\star$)}
\emph{LA-CR$\star$} represents the per-metric best performance across the two published calibration variants (LA+I, LA+S) of LA-CR~\cite{amiri2025simple}, retrained under our strict temporal split. \emph{UQ$\star$} applies the same composite convention to the values reported in~\cite{amiri2025simple} under their original dataset split, i.e., the per-metric best across the eight UQ methods they evaluated.

\mypar{Evaluation metrics}
\begin{table}[t]
\centering
\caption{\textsc{Evaluation metrics for downstream tasks}. $\uparrow$/$\downarrow$ mark whether higher/lower is better.}
\label{tab:metrics}
\setlength{\tabcolsep}{4pt}
\renewcommand{\arraystretch}{1.15}
\begin{tabular}{@{}l >{\raggedright\arraybackslash}p{2.4cm} >{\raggedright\arraybackslash}p{5.2cm} c@{}}
\toprule
\textbf{Metric} & \textbf{Task} & \textbf{Description} & \textbf{Range} \\
\midrule
\multicolumn{4}{@{}l}{\textbf{Point metrics}} \\
Acc~ & next activity            & Fraction of correct next activities        & $[0,1]\ \uparrow$ \\
MAE  & next event \& remaining time & Mean absolute error of the predicted time & $\mathbb{R}^{+}\ \downarrow$ \\
\midrule
\multicolumn{4}{@{}l}{\textbf{Uncertainty metrics}} \\
MA    & remaining time & Calibration: gap between empirical and nominal coverage & $[0,0.5]\ \downarrow$ \\
MPIW  & remaining time & Sharpness: mean prediction-interval width               & $\mathbb{R}^{+}\ \downarrow$ \\
AURG & remaining time & Sparsification: error drop when rejecting uncertain cases & $\mathbb{R}\ \uparrow$ \\
\bottomrule
\end{tabular}

\end{table}
We report two families of metrics (\autoref{tab:metrics}): point metrics for all three tasks, and uncertainty metrics that score the remaining time predictive distribution. The latter are identical to those used by Amiri et al.~\cite{amiri2025simple}.

\mypar{Execution settings}
All neural models are trained with early stopping (patience $24$), except for LA-CR, which follows its original protocol~\cite{amiri2025simple}. THP converged within at most $101$ epochs, compared to $171$ for SuTraN and $169$ for ED-LSTM; it uses a learning rate of $2\times10^{-3}$ and gradient clipping at norm $1.0$ to stabilize the mixture time head. H-uni and H-mk are fitted by maximum likelihood, with equal timestamps in inter-events perturbed by less than one second to avoid numerical instability. All neural models are trained on a single NVIDIA RTX A6000 GPU, while the Hawkes processes are estimated on a 32-vCPU Intel Xeon CPU. The Monte Carlo rollout uses $S=50$ sampled continuations per prefix; as THP defines no terminal event, the rollout horizon is set to the true number of remaining events for each test prefix. All reported figures are test set results.

\subsection{Results and Discussion}
\label{sec:eval:results}

\mypar{Next event predictions}
\autoref{tab:nextevent} compares the models on next activity and next event time prediction using the point metrics described
in~\autoref{sec:eval:setup}. H-uni, which models inter-event times without marks, yields no next activity prediction (``--''). 

For next activity prediction, SuTraN and ED-LSTM achieve the highest accuracy on $8$ of the $10$ datasets. However, THP-M is highly competitive in Group~A: it attains the best overall accuracy on \textsc{BPIC12} ($87.3\%$ vs.\ $86.1\%$ for THP-B and $84.5\%$ for ED-LSTM) and the highest accuracy among neural models on \textsc{BPIC13I} ($67.0\%$ vs.\ $50.2\%$ for ED-LSTM, the best among the state-of-the-art multi-task PPM models). Overall, THP-B and THP-M exhibit similar next activity accuracy, though notable gaps remain on certain Group~B datasets (e.g., $80.3\%$ vs.\ $71.1\%$ on \textsc{BPIC20DD}).

\begin{table*}[!ht]
\centering
\caption{\textsc{Next event predictions}.}
\label{tab:nextevent}
\small
\setlength{\tabcolsep}{1.5pt}
\renewcommand{\arraystretch}{1.1}
\providecommand{\rotcol}[1]{\rotatebox{90}{#1}}
\begin{adjustbox}{max width=\textwidth}
\begin{tabular}{l r >{\columncolor{thpmix}}r rrrr r >{\columncolor{thpmix}}r rrrr}
\toprule
& \multicolumn{6}{c}{\makecell{\textbf{Next activity}\\\textbf{accuracy (\%) $\uparrow$}}}
& \multicolumn{6}{c}{\makecell{\textbf{Next event time}\\\textbf{MAE (days) $\downarrow$}}} \\
\cmidrule(lr){2-7}\cmidrule(lr){8-13}
Log & \multicolumn{1}{c}{\rotcol{THP-B}} & \multicolumn{1}{>{\columncolor{thpmix}}c}{\rotcol{THP-M}} & \multicolumn{1}{c}{\rotcol{SuTraN}} & \multicolumn{1}{c}{\rotcol{ED-LSTM}} & \multicolumn{1}{c}{\rotcol{H-uni}} & \multicolumn{1}{c}{\rotcol{H-mk}}
        & \multicolumn{1}{c}{\rotcol{THP-B}} & \multicolumn{1}{>{\columncolor{thpmix}}c}{\rotcol{THP-M}} & \multicolumn{1}{c}{\rotcol{SuTraN}} & \multicolumn{1}{c}{\rotcol{ED-LSTM}} & \multicolumn{1}{c}{\rotcol{H-uni}} & \multicolumn{1}{c}{\rotcol{H-mk}} \\
\midrule
\multicolumn{13}{l}{\textbf{Group A}} \\
\textsc{Sepsis}   & 62.3 & 61.8 & 59.9 & \textbf{62.5} & -- & 50.6 & 0.27 & \textbf{0.26} & 1.15 & 1.10 & 0.46 & 0.46 \\
\textsc{BPIC13I}  & 61.9 & \textbf{67.0} & 48.3 & 50.2 & -- & \textbf{67.9} & \textbf{0.75} & 0.77 & 0.92 & 0.87 & 0.87 & 0.86 \\
\textsc{BPIC12}   & 86.1 & \textbf{87.3} & 77.3 & 84.5 & -- & 67.0 & 0.30 & \textbf{0.29} & 0.41 & 0.34 & 0.56 & 0.58 \\
\midrule
\multicolumn{13}{l}{\textbf{Group B}} \\
\textsc{HelpDesk}  & 56.0 & 60.3 & \textbf{73.8} & \textbf{73.8} & -- & 44.4 & 6.66 & 6.83 & 4.92 & \textbf{4.61} & 7.57 & 7.26 \\
\textsc{BPIC20DD}  & 80.3 & 71.1 & \textbf{91.1} & 88.9 & -- & 59.8 & 1.38 & 1.40 & \textbf{1.31} & 1.35 & 1.86 & 1.68 \\
\textsc{BPIC20RFP} & 82.0 & 80.8 & \textbf{89.8} & 87.2 & -- & 57.2 & 1.61 & 1.60 & \textbf{1.55} & \textbf{1.55} & 2.10 & 1.88 \\
\textsc{BPIC20PTC} & 78.2 & 77.5 & 80.9 & \textbf{87.4} & -- & 48.3 & 3.64 & 3.84 & 2.13 & \textbf{2.04} & 5.08 & 4.64 \\
\textsc{BPIC20ID}  & 80.1 & 81.2 & \textbf{90.3} & 89.7 & -- & 48.8 & 6.84 & 6.54 & \textbf{3.27} & 3.32 & 9.81 & 9.12 \\
\textsc{BPIC20TPD} & 76.0 & 69.7 & 84.0 & \textbf{84.1} & -- & 42.3 & 4.91 & 5.15 & 4.86 & \textbf{4.83} & 7.10 & 6.64 \\
\midrule
\multicolumn{13}{l}{\textbf{Outlier}} \\
\textsc{BPIC15-1}  & 6.6 & 8.9 & \textbf{16.5} & 10.6 & -- & 0.3 & 2.16 & 2.21 & \textbf{1.53} & 1.60 & 2.21 & 1.81 \\
\bottomrule
\end{tabular}
\end{adjustbox}
\end{table*}
 
Next event time prediction is more clearly group-dependent. THP-B and THP-M perform particularly well on high-frequency, small-vocabulary datasets (Group~A) --- on \textsc{SEPSIS}, THP-M reaches an MAE of $0.26$\,d, more than $4\times$ lower than that of SuTraN ($1.15$\,d) and ED-LSTM ($1.10$\,d) --- whereas SuTraN and ED-LSTM tend to achieve lower time errors on low-frequency datasets with more complex activity vocabularies (Group~B). Overall, no model consistently dominates the conventional point metrics across all datasets.

\mypar{Remaining time prediction}
\autoref{tab:rtuq} reports remaining time prediction results. Under the conventional point metric (MAE), there is no clear winner: some models achieve the lowest MAE on a similar number of datasets (e.g., 4/10 for THP-M, 3/10 for ED-LSTM). As in the next event prediction task, THP-M tends to perform better in Group~A (e.g., on \textsc{SEPSIS} it attains $9.46$\,d against $14.83$\,d for SuTraN).
\begin{table*}[!ht]
\centering
\caption{\textsc{Remaining time prediction}: the point metric RT-MAE, and the uncertainty metrics MA, MPIW, and AURG among the uncertainty-aware models (THP-M, LA-CR$\star$, UQ$\star$); best per block in bold, MoTPP columns shaded.}

\label{tab:rtuq}
\small
\setlength{\tabcolsep}{1.5pt}
\renewcommand{\arraystretch}{1.1}
\providecommand{\rotcol}[1]{\rotatebox{90}{#1}}
\begin{adjustbox}{max width=\textwidth}
\begin{tabular}{l r >{\columncolor{thpmix}}r rrr >{\columncolor{thpmix}}r rr >{\columncolor{thpmix}}r rr >{\columncolor{thpmix}}r rr}
\toprule
& \multicolumn{5}{c}{\textbf{RT-MAE (d) $\downarrow$}}
& \multicolumn{3}{c}{\textbf{MA $\downarrow$}}
& \multicolumn{3}{c}{\textbf{MPIW $\downarrow$}}
& \multicolumn{3}{c}{\textbf{AURG $\uparrow$}} \\
\cmidrule(lr){2-6}\cmidrule(lr){7-9}\cmidrule(lr){10-12}\cmidrule(lr){13-15}
Log & \multicolumn{1}{c}{\rotcol{THP-B}} & \multicolumn{1}{>{\columncolor{thpmix}}c}{\rotcol{THP-M}} & \multicolumn{1}{c}{\rotcol{SuTraN}} & \multicolumn{1}{c}{\rotcol{ED-LSTM}} & \multicolumn{1}{c}{\rotcol{UQ$\star$}}
 & \multicolumn{1}{>{\columncolor{thpmix}}c}{\rotcol{THP-M}} & \multicolumn{1}{c}{\rotcol{LA-CR$\star$}} & \multicolumn{1}{c}{\rotcol{UQ$\star$}}
 & \multicolumn{1}{>{\columncolor{thpmix}}c}{\rotcol{THP-M}} & \multicolumn{1}{c}{\rotcol{LA-CR$\star$}} & \multicolumn{1}{c}{\rotcol{UQ$\star$}}
 & \multicolumn{1}{>{\columncolor{thpmix}}c}{\rotcol{THP-M}} & \multicolumn{1}{c}{\rotcol{LA-CR$\star$}} & \multicolumn{1}{c}{\rotcol{UQ$\star$}} \\
\midrule
\multicolumn{15}{l}{\textbf{Group A}} \\
\textsc{Sepsis}   & 27.42 & \textbf{9.46} & 14.83 & 15.08 & 15.32 & \textbf{0.04} & 0.08 & 0.04 & 14.05 & \textbf{3.55} & 4.86 & 6.63 & 4.17 & \textbf{8.35} \\
\textsc{BPIC13I}  & 8.57 & 5.55 & 3.20 & 3.47 & \textbf{3.06} & \textbf{0.04} & 0.12 & 0.04 & 7.00 & \textbf{0.37} & 8.05 & \textbf{1.99} & 0.20 & 1.43 \\
\textsc{BPIC12}   & 6.76 & \textbf{3.48} & 4.69 & 7.85 & 5.86 & \textbf{0.04} & 0.06 & 0.05 & 73.85 & \textbf{2.76} & 8.63 & 1.80 & 1.26 & \textbf{1.86} \\
\midrule
\multicolumn{15}{l}{\textbf{Group B}} \\
\textsc{HelpDesk}  & 8.82 & 10.86 & 4.22 & \textbf{3.77} & 9.45 & 0.05 & \textbf{0.05} & 0.12 & 26.80 & 16.50 & \textbf{7.63} & \textbf{2.83} & 1.35 & 1.48 \\
\textsc{BPIC20DD}  & 3.56 & \textbf{3.21} & 3.25 & 3.61 & 4.12 & 0.08 & \textbf{0.02} & 0.04 & 8.16 & 3.59 & \textbf{3.37} & 1.01 & 0.47 & \textbf{2.05} \\
\textsc{BPIC20RFP} & 4.89 & \textbf{3.84} & 3.97 & 4.33 & 5.09 & 0.03 & \textbf{0.03} & 0.03 & 38.67 & \textbf{2.80} & 5.29 & 1.14 & 0.47 & \textbf{2.61} \\
\textsc{BPIC20PTC} & 12.03 & 12.65 & 6.18 & \textbf{6.11} & 7.73 & \textbf{0.04} & 0.05 & 0.05 & 65.07 & 6.94 & \textbf{5.13} & \textbf{6.59} & 5.04 & 5.71 \\
\textsc{BPIC20ID}  & 22.05 & 22.13 & 10.13 & \textbf{10.10} & 14.08 & \textbf{0.03} & 0.10 & 0.03 & 60.13 & \textbf{2.94} & 15.77 & 11.29 & \textbf{11.45} & 8.36 \\
\textsc{BPIC20TPD} & 25.61 & 22.56 & \textbf{17.60} & 18.26 & 25.02 & 0.06 & 0.06 & \textbf{0.05} & 34.89 & \textbf{10.63} & 35.57 & 10.48 & 4.41 & \textbf{14.45} \\
\midrule
\multicolumn{15}{l}{\textbf{Outlier}} \\
\textsc{BPIC15-1}  & 51.42 & 30.49 & 28.32 & 31.39 & \textbf{24.71} & \textbf{0.03} & 0.11 & 0.03 & 3276.69$^\ddagger$ & \textbf{16.85} & 20.25 & \textbf{16.56} & 7.01 & 12.61 \\
\midrule
\itshape THP-M best & & 4/10 & & & & 6/10 & & & 0/10 & & & 4/10 & & \\
\bottomrule
\end{tabular}
\end{adjustbox}
\end{table*}

Under uncertainty metrics, THP-M achieves the lowest miscalibration area (MA) on $6$ of the $10$ datasets, winning on more datasets than either UQ$\star$ or LA-CR$\star$, and the highest AURG on $4$ datasets. Its main weakness is sharpness: THP-M does not achieve the lowest MPIW on any dataset, indicating that the competing recalibrated regression methods generally produce narrower prediction intervals.

\mypar{Horizon asymmetry} The remaining time evaluation is asymmetric. THP-M relies on the oracle remaining length $J$ (\autoref{sec:inference}) for Monte Carlo rollouts and the ground-truth suffix for point predicts, inherently teacher-forcing intermediate events. In contrast, other models predict the remaining time via a dedicated head without access to this true event count. This structural advantage affects THP-M's entire remaining time block, encompassing both the point MAE and uncertainty metrics. Consequently, these results should be interpreted with this caveat. Restoring horizon parity via a self-terminating end-of-case symbol is left to future work.

\mypar{Efficiency}
\autoref{tab:timing} shows that the distributional gains cost nothing at deployment. The Total/Mean row sums training cost and averages inference cost over the $10$ logs; the final row reports mean training time per epoch. LA-CR carries no $\star$ here since both calibration variants share the same trained network and Laplace fit, incurring no additional training cost; as the fit is not epoch-based, its per-epoch entry is left as ``--''. The Hawkes column reports one timed run covering both the H-uni and H-mk models. Among the neural models, THP-M is the cheapest to train ($208$\,min vs.\ $706$\,min and $316$\,min for SuTraN and ED-LSTM, respectively, and $1541$\,min for LA-CR) and the fastest at inference: first or second on all $10$ datasets, averaging $0.030$\,ms per prefix ($52\times$ faster than SuTraN, $20\times$ than ED-LSTM, and $4.6\times$ than LA-CR). Although ED-LSTM is the cheapest per epoch, THP converges in fewer epochs (\autoref{sec:eval:setup}) and thus wins on total cost. Only the classical Hawkes models are cheaper to train in total, at far lower accuracy; their closed-form prediction runs on CPU and therefore is not benchmarked for inference. The inference figures measure each model's standard test pass per prefix: a single non-autoregressive forward pass for THP, and a full autoregressive suffix decode for SuTraN and ED-LSTM (their standard implemented inference mode). For THP-M, the remaining time distribution additionally requires the Monte Carlo rollout of \autoref{sec:train-inf}, whose cost grows linearly with $S$ and the rollout length, and is excluded from \autoref{tab:timing}.

\begin{table*}[!ht]
\centering
\caption{\textsc{Training and inference cost of all models} (bold: fastest; blue: second fastest).}
\label{tab:timing}
\small
\setlength{\tabcolsep}{1.5pt}
\renewcommand{\arraystretch}{1.1}
\providecommand{\rotcol}[1]{\rotatebox{90}{#1}}
\begin{adjustbox}{max width=\textwidth}
\begin{tabular}{l r >{\columncolor{thpmix}}r rrrr r >{\columncolor{thpmix}}r rrr}
\toprule
& \multicolumn{6}{c}{\textbf{Training (min) $\downarrow$}}
& \multicolumn{5}{c}{\textbf{Inference (ms/prefix) $\downarrow$}} \\
\cmidrule(lr){2-7}\cmidrule(lr){8-12}
Log & \multicolumn{1}{c}{\rotcol{THP-B}} & \multicolumn{1}{>{\columncolor{thpmix}}c}{\rotcol{THP-M}} & \multicolumn{1}{c}{\rotcol{SuTraN}} & \multicolumn{1}{c}{\rotcol{ED-LSTM}} & \multicolumn{1}{c}{\rotcol{Hawkes}} & \multicolumn{1}{c}{\rotcol{LA-CR}}
        & \multicolumn{1}{c}{\rotcol{THP-B}} & \multicolumn{1}{>{\columncolor{thpmix}}c}{\rotcol{THP-M}} & \multicolumn{1}{c}{\rotcol{SuTraN}} & \multicolumn{1}{c}{\rotcol{ED-LSTM}} & \multicolumn{1}{c}{\rotcol{LA-CR}} \\
\midrule
\multicolumn{12}{l}{\textbf{Group A}} \\
\textsc{Sepsis}   & 17.62 & 5.93 & 27.08 & \textcolor{nbest}{3.28} & \textbf{0.97} & 53.40 & \textbf{0.02} & \textbf{0.02} & 1.40 & 0.63 & \textcolor{nbest}{0.32} \\
\textsc{BPIC13I}  & 15.65 & \textcolor{nbest}{4.75} & 41.92 & 25.53 & \textbf{0.50} & 5.90 & \textbf{0.02} & \textbf{0.02} & 0.92 & 0.37 & \textcolor{nbest}{0.15} \\
\textsc{BPIC12}   & 401.24 & \textcolor{nbest}{82.70} & 239.40 & 179.93 & \textbf{15.89} & 1090.80 & \textbf{0.01} & \textbf{0.01} & 4.99 & 1.65 & \textcolor{nbest}{0.23} \\
\midrule
\multicolumn{12}{l}{\textbf{Group B}} \\
\textsc{HelpDesk}  & 14.70 & 15.39 & 8.22 & \textcolor{nbest}{2.27} & \textbf{1.32} & 8.10 & 0.07 & \textcolor{nbest}{0.07} & 0.25 & 0.13 & \textbf{0.06} \\
\textsc{BPIC20DD}  & 18.02 & 14.85 & 44.27 & \textcolor{nbest}{5.72} & \textbf{2.28} & 32.70 & \textbf{0.05} & \textbf{0.05} & 0.17 & 0.09 & \textcolor{nbest}{0.07} \\
\textsc{BPIC20RFP} & 16.03 & 13.19 & 22.55 & \textcolor{nbest}{2.83} & \textbf{1.73} & 17.80 & \textcolor{nbest}{0.06} & \textbf{0.05} & 0.22 & 0.11 & 0.07 \\
\textsc{BPIC20PTC} & 5.07 & 7.49 & 9.10 & \textcolor{nbest}{3.05} & \textbf{0.78} & 15.70 & \textbf{0.04} & \textcolor{nbest}{0.04} & 0.34 & 0.17 & 0.06 \\
\textsc{BPIC20ID}  & 15.31 & 14.41 & 31.45 & \textcolor{nbest}{5.25} & \textbf{2.12} & 38.80 & \textbf{0.02} & \textbf{0.02} & 0.37 & 0.17 & \textcolor{nbest}{0.07} \\
\textsc{BPIC20TPD} & 27.00 & 26.01 & 84.74 & \textcolor{nbest}{15.89} & \textbf{3.73} & 125.30 & \textbf{0.02} & \textbf{0.02} & 0.70 & 0.32 & \textcolor{nbest}{0.18} \\
\midrule
\multicolumn{12}{l}{\textbf{Outlier}} \\
\textsc{BPIC15-1}  & 38.56 & \textcolor{nbest}{23.73} & 197.35 & 72.92 & \textbf{3.84} & 153.40 & \textbf{0.01} & \textcolor{nbest}{0.01} & 6.28 & 2.41 & 0.18 \\
\midrule
\textbf{Total / Mean} & 569.20 & \textcolor{nbest}{208.45} & 706.08 & 316.67 & \textbf{33.15} & 1541.90 & \textbf{0.03} & \textbf{0.03} & 1.56 & 0.60 & \textcolor{nbest}{0.14} \\
Mean (min/epoch) & 0.64 & \textcolor{nbest}{0.59} & 0.89 & \textbf{0.26} & 3.32 & -- & & & & & \\
\bottomrule
\end{tabular}
\end{adjustbox}
\end{table*}

Overall, the results reveal a trade-off rather than uniform dominance. THP-M is competitive on conventional point metrics, particularly for high-frequency logs, while its clearest advantages lie in remaining time calibration and inference efficiency. At the same time, discriminative sequence models remain stronger on next activity prediction for several larger-vocabulary logs, and recalibrated regression methods produce sharper remaining time intervals.

\section{Related Work}
\label{sec:rw}
PPM can be seen as a form of \emph{predictive service monitoring} in service-oriented computing: it forecasts the behavior of running services or process executions to enable proactive adaptation, such as anticipating SLA or deadline violations before they occur~\cite{leitner2010runtime,metzger2015ComparingAC}. Whereas these works predict service performance with discriminative and deterministic models, we approach PPM with a generative and probabilistic model: a temporal point process.
We relate our work to three research directions in PPM: sequence models, GNNs, and TPPs for event prediction.

\mypar{Sequence models for PPM}
A comprehensive benchmark found that no single architecture dominates
across logs, with RNNs generally outperforming CNNs~\cite{rama2021deep}; more
recent sequence models are predominantly
Transformer-based~\cite{bukhsh2021processtransformer,wuyts2024sutran}.
These models reach state-of-the-art accuracy, but are discriminative and
sequence-driven: they operate on discrete steps and treat time as an auxiliary
feature~\cite{nguyen2020time,lischka2025directly} rather than modeling its
continuous dynamics.

\mypar{GNNs for PPM}
GNN-based methods instead capture the structural, non-linear dependencies in
event logs~\cite{amiri2024pgtnet,weinzierl2021exploring,hennig2025temporal};
PGTNet~\cite{amiri2024pgtnet}, for instance, combines graph representations with
temporal features for remaining time prediction. 
Recently, the focus of GNNs has shifted towards discovering more complex dependencies in event logs, rather than developing uncertainty-aware algorithms; we therefore omit GNNs from our evaluation.

\mypar{TPPs for event prediction}
TPPs are an established generative paradigm for event sequences in continuous
time, from the classical Hawkes process~\cite{hawkes1971point} to neural variants~\cite{du2016recurrent,mei2017neural}
such as the THP~\cite{zuo2020transformer}. Their use
for PPM, however, has seen only a single preliminary
attempt~\cite{nguyen2024temporal}. To our knowledge, we are the first to adapt a
neural TPP to multi-task PPM and to handle the equal timestamps pervasive in real
service execution logs, bringing generative, uncertainty-aware prediction to
predictive service monitoring.

\section{Conclusion}
\label{sec:conclusion}
We investigated temporal point processes (TPPs) as a generative paradigm for
multi-task PPM and proposed \textit{MoTPP}, which extends the Transformer
Hawkes Process to natively handle the equal timestamps of
real event logs while maintaining joint prediction of 
the next activity and its time. On ten real-world logs, MoTPP is competitive---though not
dominant---under conventional point metrics, yet it is the best-calibrated
remaining time predictor among the evaluated methods and, among the neural models,
the cheapest to train and the fastest at inference. These properties are
precisely what runtime service monitoring requires: a calibrated distribution lets
an operator act on the \emph{risk} of an SLA or deadline breach rather than on a
bare point prediction, while single-pass inference keeps that decision within the
latency budget of an online service. The value of the generative paradigm thus
lies not in point accuracy, but in making predictive monitoring both trustworthy
and operational.

Several directions follow. Beyond richer TPP models and additional PPM tasks such
as suffix prediction, the most promising next step is to close the loop from
prediction to action: feeding MoTPP's predictive distribution into the
runtime adaptation of service compositions, so that decisions such as
SLA-preserving rerouting or admission control are driven by quantified risk rather
than point forecasts. Progress here is nonetheless bounded by data---coarse
timestamp granularity and the scarcity of diverse, high-quality event logs limit
what any model can learn---which we hope encourages the community to evaluate PPM
beyond point accuracy and to invest in richer benchmarks. Establishing TPPs as a
generative, uncertainty-aware foundation for predictive service monitoring is, we
believe, a promising step in this direction.

\section*{Acknowledgements}
\label{sec:ack}
This work benefited from extended multi-turn discussions with Claude (Fable~5
and Opus~5). The model was also used to implement the code, but not to select
the datasets, train the models, or decide which results are reported. Every
AI-generated component was verified by the authors before use.

\bibliographystyle{splncs04}
\bibliography{reference}

\end{document}